\documentclass[runningheads]{llncs}

\usepackage{eccv}

\usepackage{eccvabbrv}

\usepackage{graphicx}
\usepackage{booktabs}
\usepackage{multirow}
\usepackage{booktabs}

\usepackage{colortbl}
\definecolor{class1}{gray}{.9}
\definecolor{class2}{gray}{1}
\definecolor{class3}{gray}{.9}

\usepackage[accsupp]{axessibility}  

\usepackage{hyperref}

\usepackage{orcidlink}

\begin{document}

\title{An Incremental Unified Framework for Small Defect Inspection} 


\author{Jiaqi Tang\inst{1,2,3}
\and
Hao Lu\inst{1,2}
\and
Xiaogang Xu\inst{4}
\and
Ruizheng Wu\inst{5}
\and
Sixing Hu\inst{5}
\and
Tong Zhang\inst{6}
\and
Tsz Wa Cheng\inst{6}
\and
Ming Ge\inst{7}
\and
Ying-Cong Chen\inst{1,2,3}\thanks{Corresponding author.}
\and
Fugee Tsung\inst{1,2}
}

\authorrunning{J.~Tang et al.}


\institute{Hong Kong University of Science and Technology (Guangzhou) \and
Hong Kong University of Science and Technology \and
HKUST(GZ) -- SmartMore Joint Lab \and Chinese University of Hong Kong \and
SmartMore Corporation \and
Hong Kong Industrial Artificial Intelligence \& Robotics Centre \and
Hong Kong Productivity Council
\\
\email{\{jtang092, hlu585\}@connect.hkust-gz.edu.cn} \\
\email{xiaogangxu00@gmail.com} \ \
\email{\{ruizheng.wu, david.hu\}@smartmore.com} \\
\email{\{jacquelinezhang, alancheng\}@hkflair.org} \ \
\email{mingge@hkpc.org} \\
\email{yingcongchen@hkust-gz.edu.cn} \ \
\email{season@ust.hk}
}

\maketitle
\begin{abstract}
Artificial Intelligence (AI)-driven defect inspection is pivotal in industrial manufacturing. However, existing inspection systems are typically designed for specific industrial products and struggle with diverse product portfolios and evolving processes. Although some previous studies attempt to address object dynamics by storing embeddings in the reserved memory bank, these methods suffer from memory capacity limitations and object distribution conflicts. To tackle these issues, we propose the Incremental Unified Framework (IUF), which integrates incremental learning into a unified reconstruction-based detection method, thus eliminating the need for feature storage in the memory. Based on IUF, we introduce Object-Aware Self-Attention (OASA) to delineate distinct semantic boundaries. We also integrate Semantic Compression Loss (SCL) to optimize non-primary semantic space, enhancing network adaptability for new objects. Additionally, we prioritize retaining the features of established objects during weight updates. Demonstrating prowess in both image and pixel-level defect inspection, our approach achieves state-of-the-art performance, supporting dynamic and scalable industrial inspections. \\Our code is released at \url{https://github.com/jqtangust/IUF}.
\keywords{Small Defect Inspection; Incremental Unified Framework}
\end{abstract}

\section{Introduction}

AI-driven small defect inspection, commonly known as anomaly detection, is pivotal in numerous industrial manufacturing sectors~\cite{huong2022federated,lu2022deep,zhou2020siamese,yang2024defectspectrumgranularlook}, spanning from medical engineering~\cite{khalil2022efficient} to material science~\cite{kahler2022anomaly}, and electronic components~\cite{jayasekara2023detecting}. Automated product inspections facilitated by these applications not only boast impressive accuracy but also significantly reduce labor costs.

\begin{figure*}[t]
\centering
\includegraphics[width=1.0\textwidth]{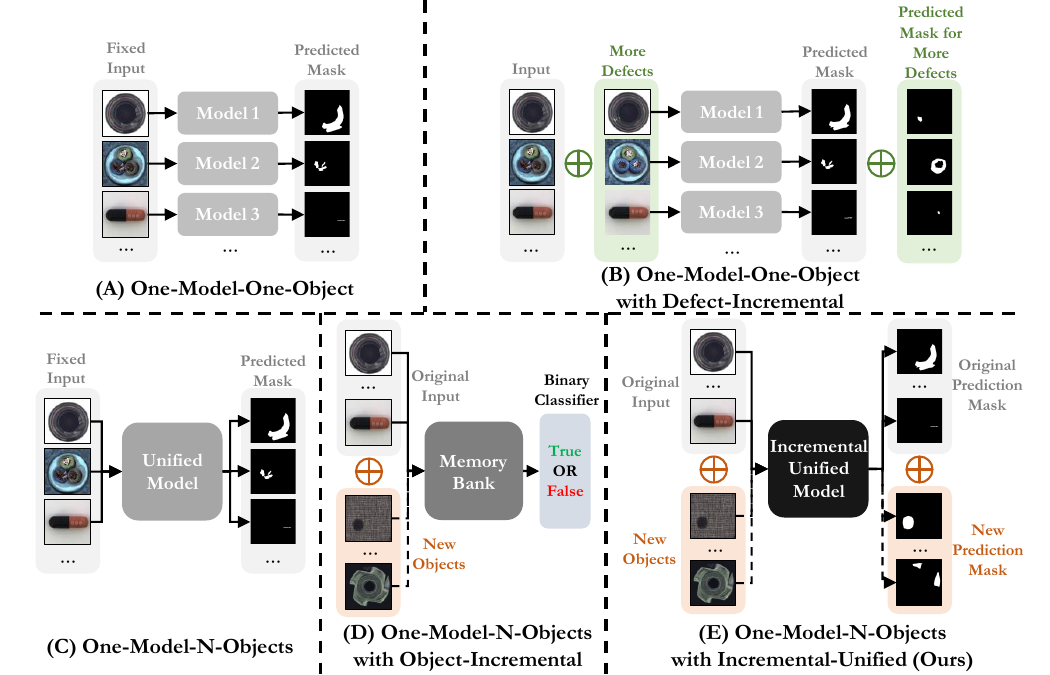}
\caption{Different framework in small defect inspection. (A) shows the most common One-Model-One-Object pattern~\cite{defard2021padim,li2022towards,liu2020towards,cohen2020sub,reiss2021panda}, which trains a separate model for each of the different objects. (B), based on (A), the types of defects are incrementing~\cite{yildiz2022automated,chang2022tire,sun2023continual,sun2022new,chen2021defect}, which improves the generalization performance in detecting different defects. (C) shows a unified model~\cite{you2022unified,zhao2023omnial} for multi-objects. (D) use a memory bank to incrementally record features for all objects for distinguishing~\cite{li2022towards}. (E) is our Incremental Unified framework, and it combines the advantages of both (C) and (D).}
\label{moti}
\end{figure*}

While most of the current inspection systems~\cite{defard2021padim,li2022towards,liu2020towards,cohen2020sub,reiss2021panda,yildiz2022automated,chang2022tire,sun2023continual,sun2022new,chen2021defect} are specifically designed for particular industrial products (One-Model-One-Object, cf. Fig.~\ref{moti} (A) (B)), the ever-changing dynamics of real-world product variants present two salient challenges. Firstly, there is a pressing need for a system capable of detecting multiple objects. Secondly, the adaptability of these systems to the frequently adjusted production schedules~\cite{towill1997analysis} is a lingering concern. Relying solely on acquiring new inspection equipment is neither cost-effective nor efficient due to increased deployment time.

\begin{figure}[t]
\centering
\includegraphics[width=0.7\linewidth]{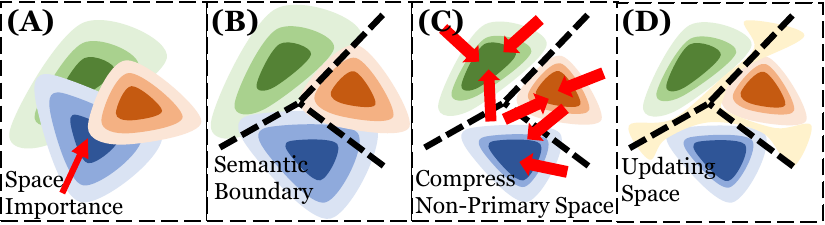}
\caption{
Semantic space in our methodology (B-D). In (A), all objects in the original semantic space are tightly coupled, leading to ``catastrophic forgetting'', i.e., the learning of new objects will result in forgetting previously learned objects. Our method firstly builds a semantic boundary of each object (B), then compacts the non-primary semantic space (C), and finally suppresses semantic updating in the previous objects' feature space (D).
}
\label{motimethod}
\end{figure}

Currently, only limited research addresses these challenges. One-Model-N-Object approaches like UniAD~\cite{you2022unified} and OmiAD~\cite{zhao2023omnial} propose a unified network suitable for a broad spectrum of objects (Fig.~\ref{moti} (C)). However, its capability is restricted when encountering dynamic or unfamiliar objects. On the other hand, CAD~\cite{li2022towards} suggests embedding features of multiple objects into a memory bank, with a binary classifier discerning defects (Fig.~\ref{moti} (D)). Yet, the differences in object distributions can jeopardize memory bank features and impede pixel-level performance evaluation. Although Liu et al.~\cite{liu2024unsupervised} design a learnable prompt to find needed object information in the memory bank, this solution is still constrained by escalating storage demands.

To surmount the confines of prior research, we introduce an Incremental Unified Framework (IUF), which integrates the multi-object unified model with object-incremental learning, as depicted in Fig.~\ref{moti} (E). IUF capitalizes on the unified model's prowess, enabling pixel-precise defect inspection across diverse objects without necessitating an embedded feature storage in the memory bank.

In this framework, one significant challenge is ``catastrophic forgetting'', which hampers network efficacy in the Incremental Unified framework due to semantic feature conflicts in the reconstruction network (Fig.~\ref{motimethod} (A)). We propose Object-Aware Self-Attention (OASA), leveraging object category features to segregate the semantic spaces of different objects (Fig.\ref{motimethod} (B)). These features serve as semantic constraints within the encoding process of the reconstruction network, establishing distinct feature boundaries in the semantic hyperplane, thereby alleviating feature coupling.  Then, we introduce Semantic Compression Loss (SCL) to condense non-essential feature values, concentrating the feature space on principal components (Fig.\ref{motimethod} (C)). This creates more semantic space for optimizing unseen objects and minimizing potential forgetting issues. Finally, when learning novel objects, we propose a new updating strategy to retain features of prior objects and reduce interference from new objects in the prevailing feature space (Fig.\ref{motimethod} (D)).
Our framework offers a versatile solution for small defect inspection, adeptly navigating the challenges posed by feature space conflicts, and achieves state-of-the-art (SOTA) performance at both image and pixel levels.

To summarize, our contributions are listed as follows:

\begin{itemize}
    \item We propose the Incremental Unified Framework (IUF), which can dynamically detect small defects in frequently-adjusted industrial scenarios. To the best of our knowledge, IUF is the first framework to integrate incremental learning into the unified reconstruction-based small defect detection method, thereby overcoming memory bank capacity limitations and object distribution conflicts.
    \item Our framework incorporates Object-Aware Self-Attention (OASA), Semantic Compression Loss (SCL), and a new updating strategy, which together reduce the potential risk of feature conflicts and mitigate the catastrophic forgetting issue.
    \item The Incremental Unified Framework (IUF) enables our method to deliver not only image-level performance but also pixel-level location. Compared to other baselines, our method achieves SOTA performance.
\end{itemize}

\section{Related Work}

\noindent \textbf{Small Defect Inspection}
Small defect inspection (Anomaly Detection) primarily falls into two paradigms: feature embedding-based and reconstruction-based methods. Both aim to pinpoint the locations of defects.

Feature embedding-based methods localize defective regions by discerning feature distribution discrepancies between defective and non-defective images, e.g., SPADE~\cite{cohen2020sub}, PaDim~\cite{defard2021padim}, PatchCore~\cite{roth2022towards}, GraphCore~\cite{xie2023pushing}, SimpleNet~\cite{liu2023simplenet}. These approaches extract diverse image features using a deep network and subsequently assess the dissimilarities between test image features and those of the training dataset. If these disparities are substantial, they indicate defective regions.
Another efficient method is the reconstruction-based approach, which relies on an alternative hypothesis. Such methods posit that models trained on normal samples excel in reproducing normal regions while struggling with anomalous regions~\cite{bergmann2018improving,chen2022utrad,you2022unified,zhao2023omnial}.
Therefore, many approaches aim to discover an optimal network architecture for effective sample reconstruction, including Autoencoder (AE)~\cite{bergmann2018improving}, Variational Autoencoder (VAE)~\cite{liu2020towards}, Generative Adversarial Networks (GANs)~\cite{akcay2019ganomaly}, Transformer~\cite{you2022unified}, etc. 
Furthermore, beyond image-level reconstruction, certain methods endeavor to reconstruct features at a more granular level, e.g., InTra~\cite{pirnay2022inpainting} and UniAD~\cite{you2022unified}.
Until recently, the majority of prior approaches adhered to the One-Model-One-Object pattern, as shown in Fig.~\ref{moti} (A). However, recent advancements such as UniAD~\cite{you2022unified} and OmniAL~\cite{zhao2023omnial} have extended the reconstruction paradigm to the One-Model-N-Objects pattern (depicted in Fig.~\ref{moti} (C)). This progress curtails memory usage compared to the one-model-one-object approach, propelling the entire problem into a more general framework.


In the context of industrial defect inspection, scenarios often evolve over time. Initial training may not encompass all objects, but rather the acquisition of objects unfolds progressively in response to manufacturing schedules and adaptations. Hence, an ideal approach would involve continual learning of new objects within a dynamic environment, while preserving the existing objects. This approach aligns with industrial evolution needs. To address this, we introduce a framework for small defect inspection under Object-Incremental Learning, as illustrated in Fig.~\ref{moti} (E).

\noindent \textbf{Incremental Learning for Small Defect Inspection}
There have been some previous studies combining incremental learning and small defect inspection; however, these studies mainly focused on augmenting the defect types of one object under a specific scenario~\cite{yildiz2022automated,chang2022tire,sun2023continual,sun2022new,chen2021defect} (Fig.~\ref{moti} (C)). The primary goal of these methods is to enhance the model's performance, which does not align with the requirements of object-incremental learning.

Recently, Li et al.\cite{li2022towards} initially proposed a feature embedding-based method for continuous learning of task sequences. However, due to the algorithm using a single memory bank to store all object information, which leads to interference between objects, the reference feature space cannot be exactly the same as the detected objects. Consequently, this algorithm cannot compute pixel-level performance and can only provide image-level performance. Most recently, Liu et al.\cite{liu2024unsupervised} proposed an unsupervised continual anomaly detection (UCAD) method. Although it attempts to distinguish features of different objects using contrastive learning, this strategy still requires the assistance of a memory feature bank.

Unlike previous solutions, our method avoids interference from multiple object classes caused by explicit memory objects. We propose the first reconstruction-based model for this continuous learning problem.
Therefore, our method can effectively avoid the problem of feature space distribution differences between different objects and only requires a reasonable feature space reassignment in the reconstruction network.




\begin{figure}[!t]
\centering
\includegraphics[width=1\linewidth]{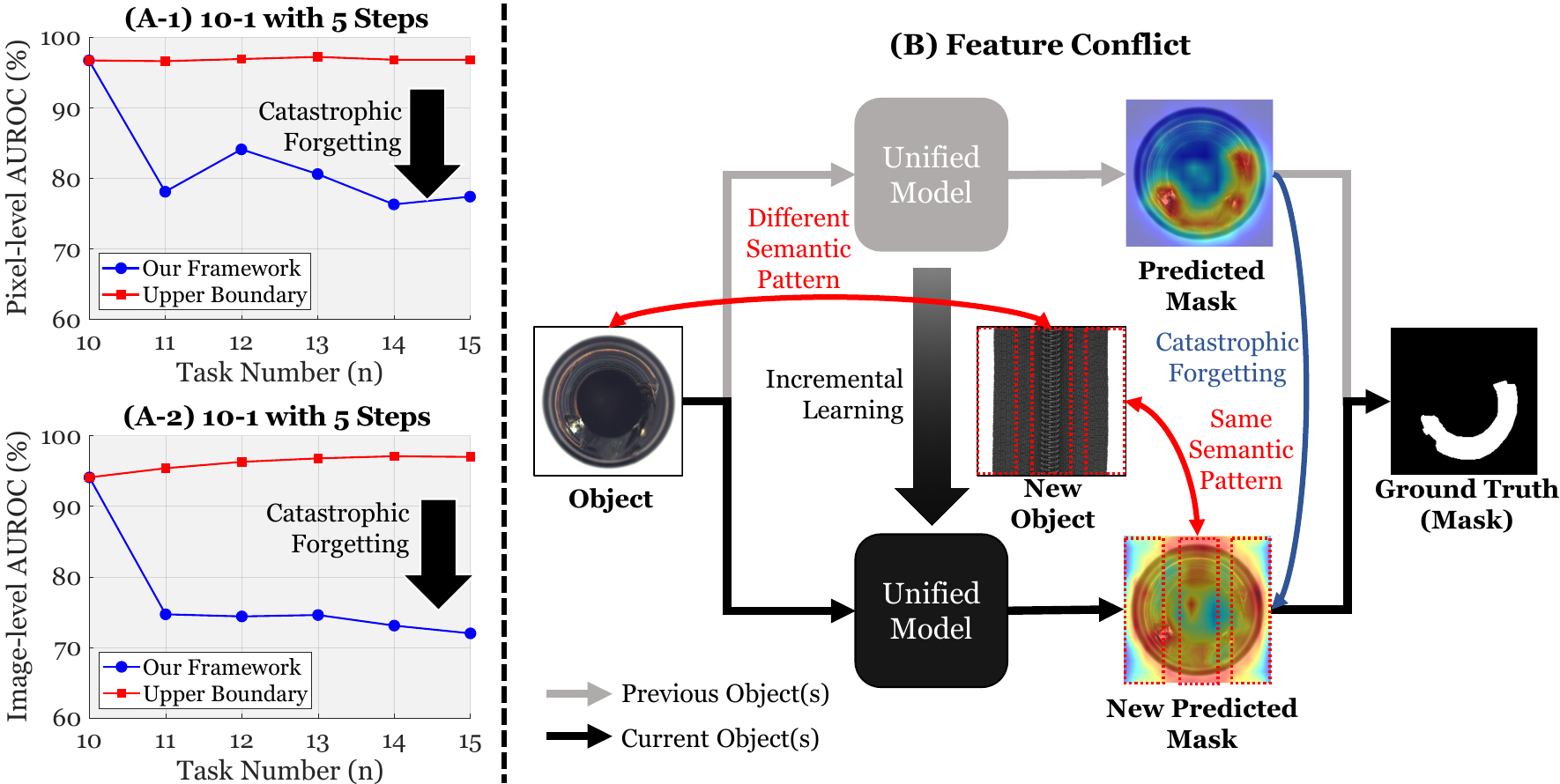}
\caption{Problems in task stream. $\mathbf{10-1\ with\ 5\ Steps}$ is an example of a task stream protocol, where we first train on 10 basic objects and then add one object at a time, with the process being completed in 5 steps (Please see Sec.~\ref{experimentsetup}(Task Protocol) for more details.). (A-1) and (A-2) demonstrate the performance of image-level and pixel-level models under the previous unified framework, UniAD~\cite{you2022unified}, where catastrophic forgetting significantly occurs. The upper boundary represents the best performance when we can use all previous objects for joint training. (B) demonstrates the reason for catastrophic forgetting. When training the current step, the training model overwrites the previous semantic patterns, causing severe feature conflicts in the reconstructed network.} 
\label{problems}
\end{figure}

\section{Problem Formulation} \label{problem}


In our Incremental Unified framework (IUF), we partition distinct objects $x$ into independent steps indexed by $t = 1 ... N$, corresponding to the number of objects. The incremental flow of involving objects into IUF can be succinctly described as $T_{n}, n \in {1, ..., N}$. During the learning process in the $t$-th step, the network undergoes updates using the model that does not access training data from previous steps ranging from 1 to $t-1$.
Among these steps, we identify normal images as $x^{+} = \{x^{+}_{o_1},x^{+}_{o_2}, \cdot \cdot \cdot, x^{+}_{o_N}\}$, while defective images are represented as $x^{-} = \{x^{-}_{o_1},x^{-}_{o_2}, \cdot \cdot \cdot, x^{-}_{o_N}\}$. Here, $o_n$ denotes distinct objects within the dataset.


For small defect inspection by reconstruction methods~\cite{you2022unified}, an autoencoder is required to reconstruct normal features in all objects, as Eq.~\ref{recon},
\begin{equation} \label{recon}
\begin{aligned}
      &\hat{x} = f_r(x^{+}; \theta), \\
    &\min L_1(\hat{x},x^+) = \min |\hat{x} - x^{+}|,
\end{aligned}
\end{equation}
where $\hat{x}$ is the reconstructed features, $\theta$ is the network parameter, and $f_r(\cdot)$ is the reconstruction network. Combining Eq.~\eqref{recon} and our framework, this process is redefined as:
\begin{equation} \label{mtrecon}
\begin{aligned}
    \textbf{Step n:}\ &\hat{x_{o_n}} = f_r(x^{+}_{o_n}; \theta), \\
    &\min L_1(\hat{x_{o_n}},x^{+}_{o_n}) = \min |\hat{x_{o_n}} - x^{+}_{o_n}|,
\end{aligned}
\end{equation}
where $n \in [ 1,N ]$.

In such task pipelines, we notice that small defect inspection is also notably susceptible to catastrophic forgetting, as shown in Fig.~\ref{problems} (A-1) and (A-2). Unlike other incremental learning tasks, small defect inspection requires the reconstruction of normal features, necessitating the decoding of semantic information from multiple channels. Without the rehearsal of the old category, the features of the new category will modify and occupy the original feature channel, resulting in semantic conflicts in space, as shown in Fig.~\ref{problems}.

Intuitively, our objective is to reorganize distinct categories of features into their respective channels, while avoiding any potential semantic conflicts. Nevertheless, within our framework, the task of establishing meaningful semantic relationships between these diverse categories presents a considerable challenge.

\section{Methodology} \label{methodology}


\noindent \textbf{Overview} To mitigate catastrophic forgetting, we propose three critical components for Incremental Unified small defect inspection. Firstly, we identify the semantic boundary to highlight the differences among categories, thereby facilitating the preservation of old knowledge (Sec.~\ref{m1}). Secondly, we compact the semantic space to maximize the compression of redundant features, which helps to alleviate the issue of semantic conflicts (Sec.~\ref{m2}). Finally, we sustain primary semantic memory to minimize the feature forgetting of the old object, thereby improving the retention of critical knowledge (Sec.~\ref{m3}).


\subsection{Identifying Semantics Boundary} \label{m1} 
\begin{figure*}[t]
\centering
\includegraphics[width=1.0\textwidth]{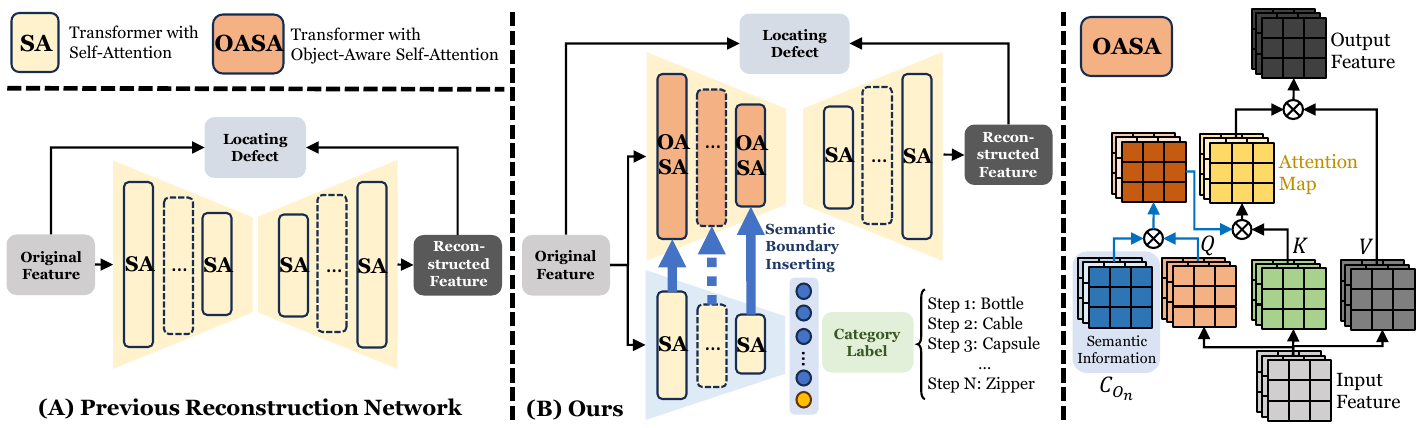}
\caption{Identify object semantics by Object-Aware Self-Attention (Sec.~\ref{m1}). (A) is the reconstructed network from the previous method~\cite{you2022unified}. (B) is our current setup, which inserts the category attributes of an image into the reconstructed network via Object-Aware Self-Attention, thus constraining the semantic space to the corresponding image features and constructing the semantic boundaries of the network.}
\label{method1}

\end{figure*}
Previous unified method~\cite{you2022unified} directly reconstructs all objects through one autoencoder, as in Fig.~\ref{method1} (A), so multiple objects share the same semantic space in one model. So, the semantic spaces of different objects are tightly coupled.

This tight semantic coupling is likely to cause undifferentiated updating of the feature space of the old objects when learning a new object, consequently resulting in a significant catastrophic forgetting of the old knowledge. Therefore, we introduce a more explicit constraint to facilitate the network to distinguish the feature semantic space among different objects. Specifically, for step $n$, we introduce its category label, $L_n$, as in Eq.~\eqref{m1-3},
\begin{equation} \label{m1-3}
\begin{aligned}
\textbf{Step n:}\ & \{y_n, C_{o_n} \}= D(x^{+}_{o_n}; \sigma), \\
&\min L_{CE}(y_n, L_n) = \min -\sum_{i=1}^{N}L_{n}^{i}\log(y_{n}^{i}),
\end{aligned}
\end{equation}
where $x^{+}_{o_n} \in \mathbb{R}^{3\times H_o\times W_o}$ indicates the normal object images in current step, $y_n$ is the final output of the discriminator, $D(\cdot)$, and $\sigma$ is the parameter in this discriminator. $L_{CE}(\cdot)$ is the cross-entropy loss, which contains $N$ objects. Besides, $D(\cdot)$ also outputs $C_{o_n} \in \mathbb{R}^{T \times C\times H\times W}$, which corresponding to $T$ key layers of semantic features, as Fig.~\ref{method1} (B). By introducing the category label, each object is limited to its corresponding semantic space in the reconstruction network, which helps to identify the semantic boundary between different objects. 

Subsequently, to introduce this semantic boundary in the reconstruction network, we designed an Object-Aware Self-Attention (OSOA) mechanism as Eq.~\eqref{m1-2},
\begin{equation} \label{m1-2}
    \text{Attention}(C_{o_n},Q,K,V)=\textnormal{softmax}(\frac{(C_{o_n} \cdot Q)  K^T}{\sqrt{d_k}})V,
\end{equation}
where $\cdot$ is Hadamard product, $d_k$ is the dimension of the key or query, and $T$ is the matrix transpose operator. $Q \in \mathbb{R}^{T \times C\times H\times W}$ is the query in $T$ key layers of the transformer-based network. $K$ and $V$ are the keys and values of this network, respectively. By inserting $C_{o_n}$ to $Q$, the reconstruction network can explicitly identify semantic boundaries.

\subsection{Compacting Semantic Space}  \label{m2}

By identifying semantic boundaries, we can further compress the feature space of old objects as much as possible, thus reserving more network capacity for new objects, which can reduce feature conflicts. To meet this goal, we design a compact feature regularization, which helps us to eliminate the redundant features of old objects and leave more capacity for future unknown new objects.


Specifically, given a training sample, the latent features $M \in \mathbb{R}^{B \times C \times H \times W}$ is firstly compressed by aggregating spatial features as in Eq.~\eqref{m2-1},
\begin{equation} \label{m2-1}
\hat{M} = \frac{1}{H \times W} \sum_{h=1}^{H} \sum_{w=1}^{W} M_n(b, c, h, w),
\end{equation} 
where $\hat{M}\in \mathbb{R}^{B \times C}$ represents the semantic information matrix of latent features in each batch. As is common in the field~\cite{Ding_2021_ICCV,9340601}, we hypothesize that different objects of semantic features are distributed on distinct channels, while spatial information is relatively irrelevant. Then, we perform SVD on the first two dimensions $B$ and $C$ of $\hat{M}$. The representation formula is as Eq.~\eqref{m2-2},
\begin{equation}
\begin{aligned}\label{m2-2}
	\hat{M}&=USV^{T}\\
	&=\underset{\mathbf{Batch\ Space}}{\underbrace{\left[ \begin{matrix}
	u_{11}&		u_{12}&		u_{13}&		\cdots&		u_{1B}\\
	u_{21}&		u_{22}&		u_{23}&		\cdots&		u_{2B}\\
	u_{31}&		u_{32}&		u_{33}&		\cdots&		u_{3B}\\
	\vdots&		\vdots&		\ddots&		\ddots&		\vdots\\
	u_{B1}&		u_{B2}&		u_{B3}&		\cdots&		u_{BB}\\
\end{matrix} \right] }}\left[ \begin{matrix}
	\sigma _1&		0&		\cdots&		0\\
	0&		\sigma _2&		\cdots&		0\\
	\vdots&		\vdots&		\ddots&		\vdots\\
	0&		0&		\cdots&		\sigma _C\\
	0&		0&		\cdots&		0\\
	\vdots&		\vdots&		\ddots&		\vdots\\
	0&		0&		\cdots&		0\\
\end{matrix} \right] \underset{\mathbf{Channel\ Space}}{\underbrace{\left( \left[ \begin{matrix}
	v_{11}&		v_{12}&		\cdots&		v_{1C}\\
	v_{21}&		v_{22}&		\cdots&		v_{2C}\\
	\vdots&		\vdots&		\ddots&		\vdots\\
	v_{C1}&		v_{C2}&		\cdots&		v_{CC}\\
\end{matrix} \right] \right) ^T}},
\end{aligned}
\end{equation}
where three matrices obtained after SVD, i.e., $U$, $S$, and $V^T$. Among them, $U \in \mathbb{R}^{B \times B}$ is an orthogonal matrix for batch eigenspace, $S = \mathrm{diag}(\sigma_1,\sigma_2,...,\sigma_C) \in \mathbb{R}^{B \times C}$ is a diagonal matrix of eigenvalues, and $V^T \in \mathbb{R}^{C \times C}$ is an orthogonal matrix of channel eigenspace.

\begin{figure}[t]
\centering
\includegraphics[width=1.0\linewidth]{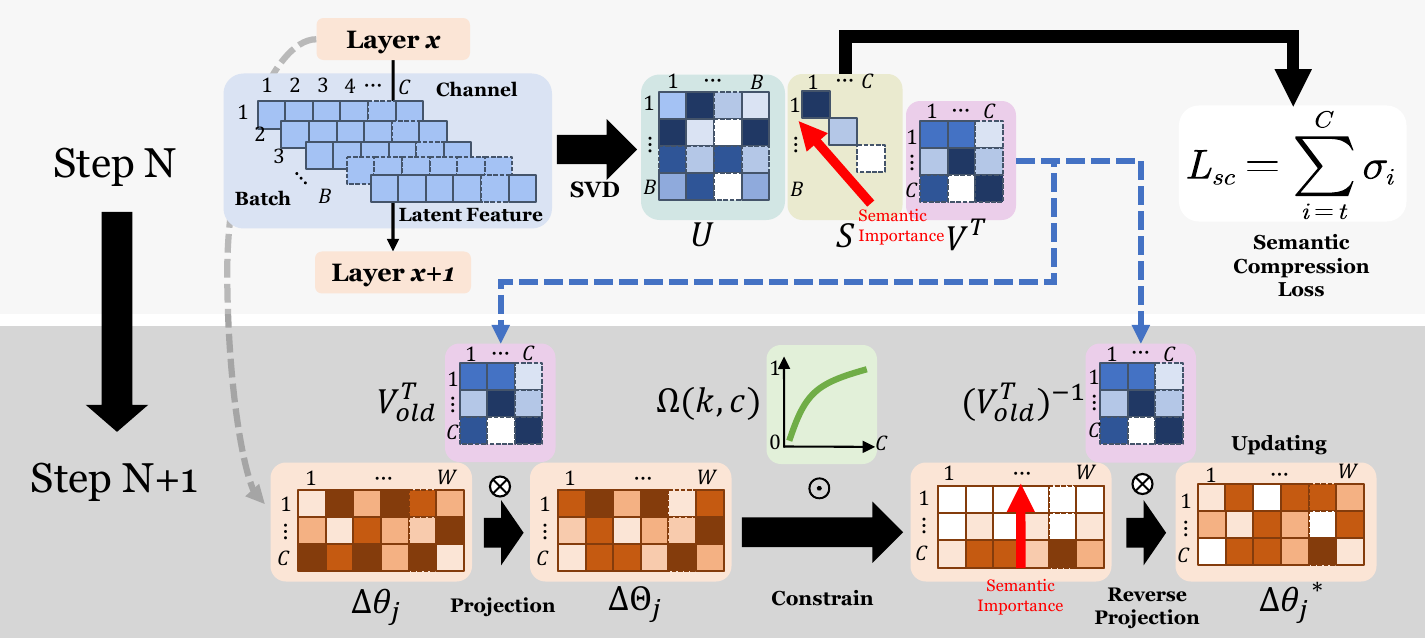}
\caption{Semantic space operation. We perform an SVD decomposition of the semantic space, based on the feature values that represent semantic importance, and compact the non-primary space in Sec.~\ref{m2}. In addition, when learning a new object, we project the update weights to the semantic space of previous features, and then block the weight updates that are semantically significant for previous information in Sec.~\ref{m3}.}
\label{method2}
\end{figure}

$S_n$ reflects the importance of semantic information for $M$ on each channel dimension. Larger eigenvalues correspond to more important semantic space, while smaller eigenvalues correspond to relatively non-primary semantic space~\cite{lin2020hrank,hoefler2021sparsity,han2023svdiff}. Based on this fact, we can compress the features of each object by reducing the non-primary semantic information. Therefore, we construct the Semantic Compression Loss (SCL) as in Eq.~\eqref{m2-3} and Fig.~\ref{method2},
\begin{equation} \label{m2-3}
    L_{sc} =   \sum_{i=t}^{C} \sigma_i ,
\end{equation}
where $t \in [1,C]$ is a hyperparameter, which is a scale factor that reflects the degree of compression of the semantic feature space. To construct the total loss, we use $L = \lambda_0 L_1 + \lambda_1 L_{CE} + \lambda_2 L_{sc}$. $\lambda_0$ is set to 1 as a standard of the base task, and $\lambda_1$ is 0.5 as the auxiliary task experience for balancing different loss values. $\lambda_2$ is within the range of $1-10$, and in practice, users can make custom adjustment of $\lambda_2$ for controlling the old objects' feature space. This loss continuously compacts the non-primary feature space while achieving the final optimization goal.

\subsection{Reinforcing Primary Semantic Memory}  \label{m3}
While we have taken measures to ensure the availability of the requisite semantic space within the network, the risk of catastrophic forgetting still persists, particularly in an unregulated task stream, when significant feature space overlap exists between older and newer objects. Hence, it becomes imperative for us to address dual concerns: firstly, how to retain the prior semantic information when updating weights, and secondly, how to simultaneously acquire knowledge about new objects without unduly influencing the semantic space associated with older objects.

Vanilla gradient descent leads to an undifferentiated update of all feature space, if $\theta = (\theta_1, \theta_2, \ldots, \theta_J) \in \mathbb{R}^{C \times W}$ is the parameter of the model, and $W$ is related to network structure. $\frac{\partial L(\theta)}{\partial \theta_j}, j = 1, 2, \ldots, J$ is the partial derivative of $\theta$. The process of model updating is shown in Eq.~\eqref{m3-1},
\begin{equation} \label{m3-1}
\begin{aligned}
     \nabla \theta_{j} &= -\alpha \frac{\partial L(\theta)}{\partial \theta_j}, \\
   \theta_{j}^{'}  &\leftarrow \theta_j + \nabla \theta_j,
\end{aligned}
\end{equation}
where $\alpha$ is the learning rate, $\nabla \theta_{j}$ is updating vector, and $\theta_{j}^{'}$ is the new weight next iteration.

\noindent \textbf{Retaining Prior Semantic Information} To maintain the semantics of the old objects, it is essential to continuously consolidate the existing weights in old objects during the incremental process. Therefore, we constantly copy the old weight in gradient descent in the next step, as in Eq.~\eqref{m3-2},
\begin{equation} \label{m3-2}
\begin{aligned}
    \theta_{j}^{'} \leftarrow \theta_j + \nabla \theta_j + \beta \theta^{old}_j,
\end{aligned}
\end{equation}
where $\theta^{old}_j$ is the old weight in previous objects, and $\beta$ is a hyper-parameter for controlling the magnitude of this regulation. 

\noindent \textbf{Decreasing Rewriting of Prior Semantics} Although we can retain prior semantic information in updating weight, the rewriting of old semantic space still happens. Therefore, suppressing weight updates in old semantic spaces will constrain the new object to use other undisturbed semantic spaces, which will further reduce feature conflicts.

Based on this motivation, we consider that the importance of the semantic space can be represented by $V^T$ (Eq.~\eqref{m2-2}). Thus, we project the updating weight, $\nabla \theta_{j}$, to the corresponding channel space in old objects, as in Eq.~\eqref{m3-3},
\begin{equation} 
\begin{aligned}\label{m3-3}
\nabla \Theta_j = V_{old}^T \nabla \theta_j,
\end{aligned}
\end{equation}
where $\nabla \Theta_j \in \mathbb{R}^{C \times W} $ is updating weight in channel space. $V_{old}^T$ is channel eigenspace in previous steps.

To constrain the updating in primary semantic space in previous steps, we empirically use a $\log$ function to constrain the updating of different channels, $c \in \mathbb{R}^{1 \times C}$, as Eq.~\eqref{m3-4},
\begin{equation} \label{m3-4}
    \Omega(k,c) =  k \times \log(c), 
\end{equation}
where $\Omega(\cdot)$ is constraining function. When the channel index $c$ equals $1$, the result of the $\log$ function is 0, indicating that the model does not update in this specific dimension. This is because the model is in the most crucial semantic space for the previous objects. As the channel index $c$ increases, this channel becomes more likely to trigger model updates, and these dimensions become less important for representing previous objects.

By acting on this function to $\nabla \Theta_j$ and then projecting it back to the original space, we finally get the updating weight in the original space, as in Eq.~\eqref{m3-5} and Fig.~\ref{method2},
\begin{equation}  \label{m3-5}
    \nabla \theta^{*}_j = (V_{old}^T)^{-1} \Omega(k,n) \odot \nabla \Theta_j,
\end{equation}
where $\odot$ is the channel-wise product. In summary, our updating is shown as Eq.~\eqref{m3-6},
\begin{equation} \label{m3-6}
\begin{aligned}
    \theta^{'}_j \leftarrow \theta_j + \nabla \theta^{*}_j + \beta \theta^{old}_j.
\end{aligned}
\end{equation}
Based on this, our model can reinforce the primary semantic memory of old objects and overcome catastrophic forgetting significantly.

\section{Experiments}

\subsection{Experimental Setup} \label{experimentsetup}
\noindent \textbf{Datasets} We choose MVTec-AD~\cite{bergmann2019mvtec} and VisA~\cite{zou2022spot} as our dataset. MVTec-AD~\cite{bergmann2019mvtec} and VisA~\cite{zou2022spot} have 15 and 12 types of objects, respectively. These two employed datasets have well-recognized complex cases for real-world evaluation, and different objects' feature domains show significantly varying distributions, so both of them can be available for our Incremental Unified Framework.



\noindent \textbf{Task Protocol} According to our framework in Fig.~\ref{moti} (E), our reconstruction model incrementally learns new objects. Based on the practical requirements of industrial defect inspection, we set up our experiments in both single-step and multi-step settings. 

We represent our task stream as $\mathbf{X - Y \ with \ N \ Step(s)}$. Here, $\mathbf{X}$ denotes the number of base objects before starting incremental learning, $\mathbf{Y}$ represents the number of new objects incremented in each step, and $\mathbf{N}$ indicates the number of tasks during incremental learning. When training on base objects, $\mathbf{N} = 0$, and after one step, $\mathbf{N} = \mathbf{N} + 1$. Our task stream is shown as follows:
\begin{itemize}
    \item MVTec-AD~\cite{bergmann2019mvtec}: $\mathbf{14-1\ with \ 1\ Step}$, $\mathbf{10-5\ with \ 1\ Step}$,\\ $\mathbf{3 - 3\ with \ 4\ Steps}$ and $\mathbf{10-1\ with \ 5\ Steps}$.
    \item VisA~\cite{zou2022spot}: $\mathbf{11-1\ with \ 1\ Step}$, $\mathbf{8-4\ with \ 1\ Step}$, $\mathbf{8-1\ with \ 4\ Steps}$.
\end{itemize}
\noindent \textbf{Baselines} We select some SOTA methods in small defect inspection as the baseline, including PaDiM~\cite{defard2021padim}, DRAEM~\cite{Zavrtanik2021DRMA}, PatchCore~\cite{roth2022towards}, PANDA~\cite{reiss2021panda}, CutPaste~\cite{Li_2021_cut}, and UniAD~\cite{you2022unified}. Notably, some baselines integrate with the incremental learning protocol, CAD~\cite{li2022towards}. In addition, we integrate several available SOTA baselines in incremental learning with UniAD, including EWC~\cite{kirkpatrick2017overcoming}, SI~\cite{zenke2017si}, MAS~\cite{aljundi2018mas} and LVT~\cite{wang2022continual}.


\begin{table*}[!t]
    \caption{Quantitative evaluation in MvTec AD~\cite{bergmann2019mvtec} (A) and VisA~\cite{zou2022spot} (B). ``Image /  \textcolor{gray}{Pixel}'' shows the image-level and pixel-level performance respectively. ``NA'' is Not Available, since CAD~\cite{li2022towards}-based method cannot locate defects in pixel-level. \textcolor{red}{Red} and \textcolor{red!50}{Gray Red} represents the best image-level and pixel-level performance respectively.}
    \centering
    (A) Quantitative Performance in MvTec~\cite{bergmann2019mvtec}.\\
    \resizebox{1.0\textwidth}{!}{
    \begin{tabular}{lcc|cc|cc|cc}
    \toprule
    \multirow{2}[4]{*}{} & \multicolumn{2}{c|}{$\mathbf{14-1\ with \ 1\ Step}$} & \multicolumn{2}{c|}{$\mathbf{10-5\ with \ 1\ Step}$} & \multicolumn{2}{c|}{$\mathbf{3 - 3\ with \ 4\ Steps}$} & \multicolumn{2}{c}{$\mathbf{10-1\ with \ 5\ Steps}$} \\
     \cmidrule{2-9}   Method       & ACC($\uparrow$)   & FM($\downarrow$)    & ACC($\uparrow$)   & FM($\downarrow$)    & ACC($\uparrow$)   & FM($\downarrow$)    & ACC($\uparrow$)   & FM($\downarrow$) \\
    \midrule
    
   \rowcolor{class1}PaDim~\cite{defard2021padim} &  57.5 / \textcolor{gray}{77.1} & 23.1  / \textcolor{gray}{20.2} & 64.4 / \textcolor{gray}{81.4}& 9.1  / \textcolor{gray}{14.1} & 60.0 / \textcolor{gray}{76.16}   & 22.6  / \textcolor{gray}{20.3}  & 53.9 / \textcolor{gray}{68.4}  & 18.1 / \textcolor{gray}{24.1} \\
   
    \rowcolor{class1}PatchCore~\cite{roth2022towards} & 66.5 / \textcolor{gray}{83.8}& 34.2 / \textcolor{gray}{24.1}&  69.6 / \textcolor{gray}{62.4}& 22.6 / \textcolor{gray}{25.2}  & 62.4 / \textcolor{gray}{77.9}  & 37.3 / \textcolor{gray}{22.1}   & 55.3 / \textcolor{gray}{73.8}  & 30.8 / \textcolor{gray}{27.5} \\
    
    \rowcolor{class1}DRAEM~\cite{Zavrtanik2021DRMA} & 51.1 / \textcolor{gray}{61.2}& 8.2  / \textcolor{gray}{8.2}& 58.0 / \textcolor{gray}{63.2}& 11.8  / \textcolor{red!60}{3.9}  & 54.9 / \textcolor{gray}{57.6}  & \textcolor{red}{2.6} / \textcolor{gray}{9.8}   & 52.3 / \textcolor{gray}{59.0}  & 13.8 / \textcolor{gray}{8.8} \\
    
    \midrule
    
    \rowcolor{class2}UniAD~\cite{you2022unified} & 85.7 / \textcolor{gray}{89.6}& 18.3 / \textcolor{gray}{13.3}& 86.7 / \textcolor{gray}{91.5}& 14.9 / \textcolor{gray}{10.6}   & 81.3 / \textcolor{gray}{88.7}  &  7.4 / \textcolor{gray}{10.6}   & 76.6 / \textcolor{gray}{82.3}  & 21.1 / \textcolor{gray}{17.3} \\
    
    \rowcolor{class2}UniAD~\cite{you2022unified} + EWC~\cite{kirkpatrick2017overcoming} & 92.8 / \textcolor{gray}{95.4}& 4.1  / \textcolor{gray}{1.9}& 90.5 / \textcolor{gray}{93.6}& 7.3  / \textcolor{gray}{4.2} &79.6  / \textcolor{gray}{89.0}  & 9.5  / \textcolor{gray}{10.1}  & 89.6 / \textcolor{gray}{93.8}  & 5.4 / \textcolor{gray}{3.6} \\
    
    \rowcolor{class2}UniAD~\cite{you2022unified} + SI~\cite{zenke2017si} & 85.7 / \textcolor{gray}{89.5}& 18.4 / \textcolor{gray}{13.4}& 84.1 / \textcolor{gray}{88.3}& 20.2 / \textcolor{gray}{17.0}   & 81.9 / \textcolor{gray}{88.5}  & 7.0  / \textcolor{gray}{10.8}  & 77.2 / \textcolor{gray}{81.6}  &  20.2 / \textcolor{gray}{18.2}  \\
    
    \rowcolor{class2}UniAD~\cite{you2022unified} + MAS~\cite{aljundi2018mas} & 85.8 / \textcolor{gray}{89.6}& 18.1 / \textcolor{gray}{13.3}&  86.8 / \textcolor{gray}{91.0}& 14.9 / \textcolor{gray}{11.6}& 81.5 / \textcolor{gray}{89.0}& 7.2  / \textcolor{gray}{10.2}& 77.9 / \textcolor{gray}{82.0}& 19.5 / \textcolor{gray}{17.7}\\
    
    \rowcolor{class2}UniAD~\cite{you2022unified} + LVT~\cite{wang2022continual} & 80.4 / \textcolor{gray}{86.0}& 29.1 / \textcolor{gray}{20.6}& 87.1 / \textcolor{gray}{90.6}& 14.1 / \textcolor{gray}{12.3}& 80.4 / \textcolor{gray}{88.6}& 8.6  / \textcolor{gray}{10.6}& 78.2 / \textcolor{gray}{88.3}& 19.1 / \textcolor{gray}{16.1}\\
    
    \midrule
    \rowcolor{class3}CAD + DNE~\cite{li2022towards}   & 84.5 / \textcolor{gray}{NA}& \textcolor{red}{-2.0} / \textcolor{gray}{NA}   & 87.8 / \textcolor{gray}{NA}& 1.1  / \textcolor{gray}{NA}  & 80.3 / \textcolor{gray}{NA}  & 6.6  / \textcolor{gray}{NA}  & 77.7 / \textcolor{gray}{NA}  & 9.7 / \textcolor{gray}{NA} \\
    
    \rowcolor{class3}CAD~\cite{li2022towards} + CutPaste~\cite{Li_2021_cut} & 84.3 / \textcolor{gray}{NA}&  -1.6 / \textcolor{gray}{NA}    & 87.1 / \textcolor{gray}{NA}  & \textcolor{red}{-0.3} / \textcolor{gray}{NA}   & 79.2 / \textcolor{gray}{NA}  & 12.6 / \textcolor{gray}{NA}   & 70.6 / \textcolor{gray}{NA}  & 20.2 / \textcolor{gray}{NA} \\
    
    \rowcolor{class3}CAD~\cite{li2022towards} + PANDA~\cite{reiss2021panda} & 50.0 / \textcolor{gray}{NA}& 6.0  / \textcolor{gray}{NA}  & 55.4 / \textcolor{gray}{NA}  & 8.9  / \textcolor{gray}{NA} & 62.4 / \textcolor{gray}{NA}  & 36.8 / \textcolor{gray}{NA}   & 51.3 / \textcolor{gray}{NA}  & 10.3 / \textcolor{gray}{NA} \\
    
    \midrule
    \rowcolor{class2}Ours  & \textcolor{red}{96.0} / \textcolor{red!60}{96.3} & 1.0  / \textcolor{red!60}{0.6}& \textcolor{red}{92.2} /  \textcolor{red!60}{94.4}& 9.3  / \textcolor{gray}{6.3} & \textcolor{red}{84.2} /  \textcolor{red!60}{91.1} & 10.0 / \textcolor{red!60}{8.4}& \textcolor{red}{94.2} / \textcolor{red!60}{95.1}& \textcolor{red}{3.2} / \textcolor{red!60}{1.0}\\
    
    \bottomrule
    \\
    \end{tabular}}

    (B) Quantitative Performance in VisA~\cite{zou2022spot}.\\
    \resizebox{1.0\textwidth}{!}{
    \small
    \begin{tabular}{lcc|cc|cc}
    \toprule
    \multirow{2}[4]{*}{} & \multicolumn{2}{c|}{$\mathbf{11-1\ with \ 1\ Step}$} & \multicolumn{2}{c|}{$\mathbf{8-4\ with \ 1\ Step}$} & \multicolumn{2}{c}{$\mathbf{8-1\ with \ 4\ Steps}$} \\
    \cmidrule{2-7}  Method        & ACC($\uparrow$)   & FM($\downarrow$)    & ACC($\uparrow$)   & FM($\downarrow$)    & ACC($\uparrow$)   & FM($\downarrow$) \\
    \midrule
    
   \rowcolor{class1}PaDim~\cite{defard2021padim} & 59.7 / \textcolor{gray}{84.3} & 20.6  / \textcolor{gray}{14.2} & 60.3 / \textcolor{gray}{84.2}& 21.8  / \textcolor{gray}{14.0} & 54.3 / \textcolor{gray}{83.4}   & 21.1  / \textcolor{gray}{9.7} \\
   
    \rowcolor{class1}PatchCore~\cite{roth2022towards} & 66.0 / \textcolor{gray}{85.6}& 30.0 / \textcolor{gray}{13.0}&  67.4 / \textcolor{gray}{86.4} & 33.3 / \textcolor{gray}{14.3}  & 56.2 / \textcolor{gray}{83.6}  & 34.8 / \textcolor{gray}{12.4}  \\
    
    \rowcolor{class1}DRAEM~\cite{Zavrtanik2021DRMA} & 48.4 / \textcolor{gray}{60.5}& 30.6  / \textcolor{gray}{15.8}& 63.6 / \textcolor{gray}{49.6}& 17.7 / \textcolor{gray}{29.7}  & 51.8 / \textcolor{gray}{63.4}  & 25.9 / \textcolor{gray}{10.5}   \\
    
    \midrule
    
    \rowcolor{class2}UniAD~\cite{you2022unified} & 75.0 / \textcolor{gray}{92.1}& 22.4 / \textcolor{gray}{11.4}& 78.1 / \textcolor{gray}{94.0}& 14.7 / \textcolor{gray}{8.4}   & 72.2 / \textcolor{gray}{90.8}  &  16.6 / \textcolor{gray}{9.2}   \\
    
    \rowcolor{class2}UniAD~\cite{you2022unified} + EWC~\cite{kirkpatrick2017overcoming} & 78.7  / \textcolor{gray}{95.4}  & 14.9  / \textcolor{gray}{4.8} & 80.5 / \textcolor{red!60}{95.4}& 10.0  / \textcolor{red!60}{5.3} &72.3  / \textcolor{gray}{92.3}  & 16.5  / \textcolor{gray}{7.3}  \\
    
    \rowcolor{class2}UniAD~\cite{you2022unified} + SI~\cite{zenke2017si} & 78.1 / \textcolor{gray}{92.0}& 16.9 / \textcolor{gray}{11.5}& \textcolor{red}{80.8} / \textcolor{gray}{93.9}& 9.2 / \textcolor{gray}{8.3}   & 69.8 / \textcolor{gray}{88.5}  & 19.8 / \textcolor{gray}{12.0}   \\
    
    \rowcolor{class2}UniAD~\cite{you2022unified} + MAS~\cite{aljundi2018mas} & 75.4 / \textcolor{gray}{91.8}& 21.5 / \textcolor{gray}{11.9}&  78.4/ \textcolor{gray}{94.0}& 14.1 / \textcolor{gray}{8.4}& 72.1 / \textcolor{gray}{90.6}& 16.7  / \textcolor{gray}{9.4}\\
    
    \rowcolor{class2}UniAD~\cite{you2022unified} + LVT~\cite{wang2022continual} & 77.5 / \textcolor{gray}{92.3}& 17.3 / \textcolor{gray}{10.9}&  78.8/ \textcolor{gray}{94.1}& 13.4 / \textcolor{gray}{8.1}& 70.8 / \textcolor{gray}{91.4}& 18.3  / \textcolor{gray}{8.4}\\
    
    \midrule
    \rowcolor{class3}CAD + DNE~\cite{li2022towards}   & 71.2 / \textcolor{gray}{NA}& \textcolor{red}{-10.2} / \textcolor{gray}{NA}   & 64.1 / \textcolor{gray}{NA}& 6.1  / \textcolor{gray}{NA}  & 58.6 / \textcolor{gray}{NA}  & 10.2 / \textcolor{gray}{NA}  \\
    
    \rowcolor{class3}CAD~\cite{li2022towards} + CutPaste~\cite{Li_2021_cut} & 65.8 / \textcolor{gray}{NA} & 3.5 / \textcolor{gray}{NA} & 63.2 / \textcolor{gray}{NA}  & 5.1 / \textcolor{gray}{NA}   & 56.2 / \textcolor{gray}{NA}  & 13.9 / \textcolor{gray}{NA} \\
    
    \rowcolor{class3}CAD~\cite{li2022towards} + PANDA~\cite{reiss2021panda} & 55.7 / \textcolor{gray}{NA}& -1.7 / \textcolor{gray}{NA}  & 56.7 / \textcolor{gray}{NA}  & \textcolor{red}{-3.3} / \textcolor{gray}{NA} & 56.0 / \textcolor{gray}{NA}  & \textcolor{red}{-0.3} / \textcolor{gray}{NA}  \\
    
    \midrule
    \rowcolor{class2}Ours  & \textcolor{red}{87.3} / \textcolor{red!60}{97.6}& 2.4  / \textcolor{red!60}{1.8}& 80.1 /  \textcolor{red!60}{95.4}& 15.2  / \textcolor{gray}{6.1} & \textcolor{red}{79.8} /  \textcolor{red!60}{95.0}& 9.8 / \textcolor{red!60}{6.8} \\
    \bottomrule
    \end{tabular}}
  \label{perfor}
\end{table*}

\begin{figure}[t]
	\centering
        \huge
	\newcommand\widthface{0.47}
	\resizebox{1.0\linewidth}{!}{
	\begin{tabular}{cccccccc}
      Input& UniAD~\cite{you2022unified} &UniAD + SI~\cite{zenke2017si} &UniAD + MAS~\cite{aljundi2018mas}&UniAD + LVT~\cite{wang2022continual}& UniAD + EWC~\cite{kirkpatrick2017overcoming} &Ours  &Ground Truth \\
	    \includegraphics[width=\widthface\textwidth]{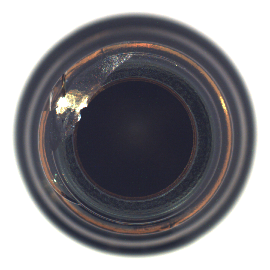} &
     	\includegraphics[width=\widthface\textwidth]{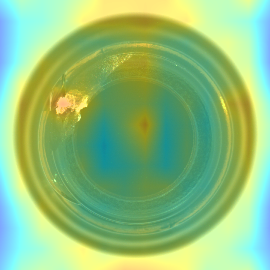} &
            \includegraphics[width=\widthface\textwidth]{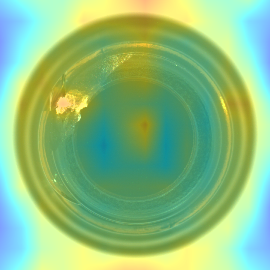} &
            \includegraphics[width=\widthface\textwidth]{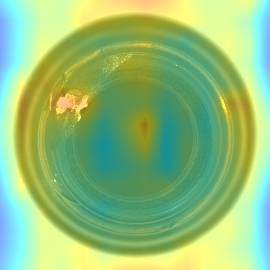} &
            \includegraphics[width=\widthface\textwidth]{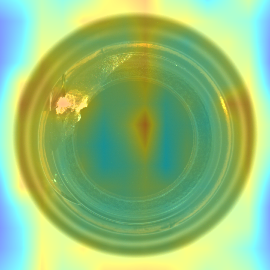} &
            \includegraphics[width=\widthface\textwidth]{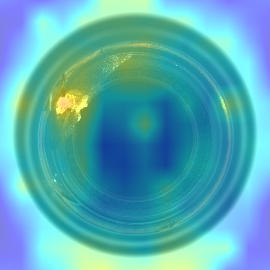} &
	    \includegraphics[width=\widthface\textwidth]{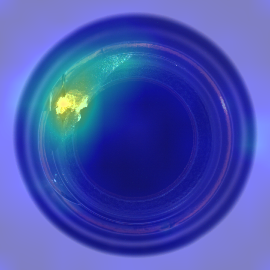} &
	    \includegraphics[width=\widthface\textwidth]{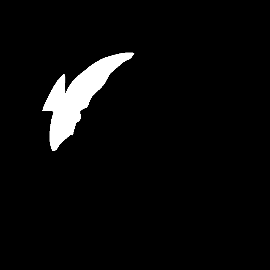} \\

	    \includegraphics[width=\widthface\textwidth]{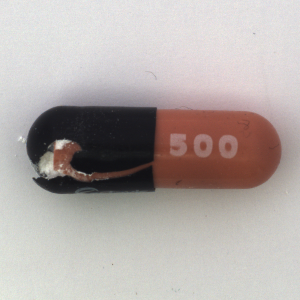} &
     	\includegraphics[width=\widthface\textwidth]{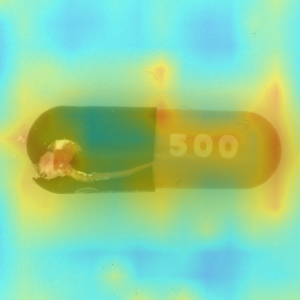} &
            \includegraphics[width=\widthface\textwidth]{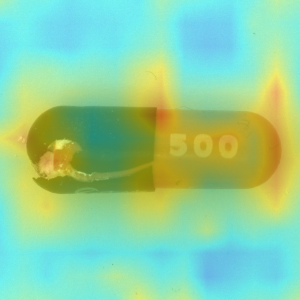} &
            \includegraphics[width=\widthface\textwidth]{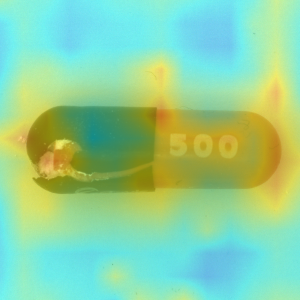} &
            \includegraphics[width=\widthface\textwidth]{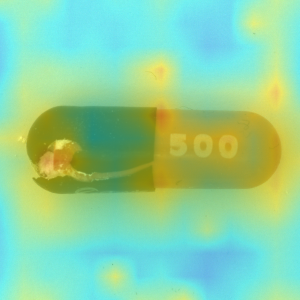} &
            \includegraphics[width=\widthface\textwidth]{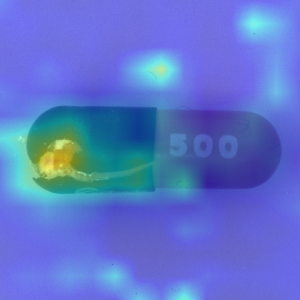} &
	    \includegraphics[width=\widthface\textwidth]{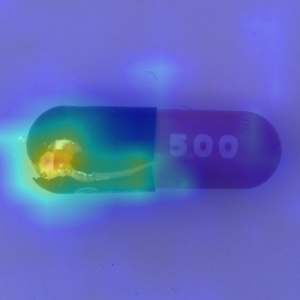} &
	    \includegraphics[width=\widthface\textwidth]{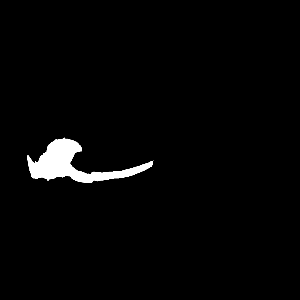} \\
     
     	\includegraphics[width=\widthface\textwidth]{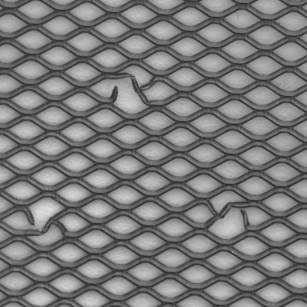} &
     	\includegraphics[width=\widthface\textwidth]{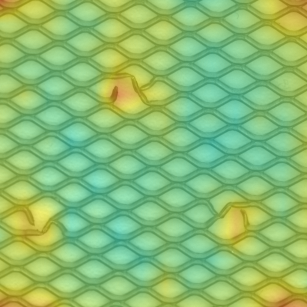} &
            \includegraphics[width=\widthface\textwidth]{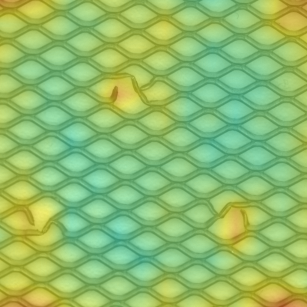} &
            \includegraphics[width=\widthface\textwidth]{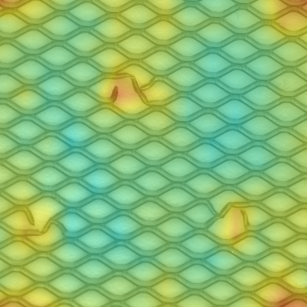} &
            \includegraphics[width=\widthface\textwidth]{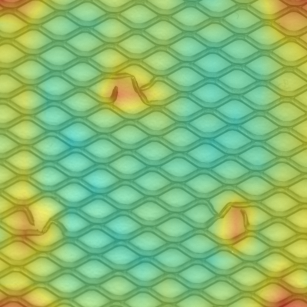} &
            \includegraphics[width=\widthface\textwidth]{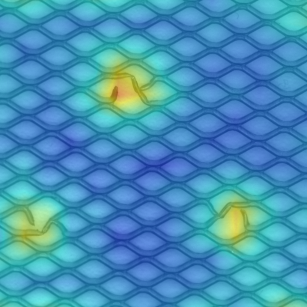} &
	    \includegraphics[width=\widthface\textwidth]{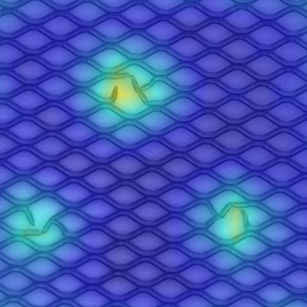} &
	    \includegraphics[width=\widthface\textwidth]{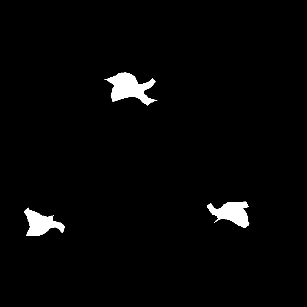} \\
     
     	\includegraphics[width=\widthface\textwidth]{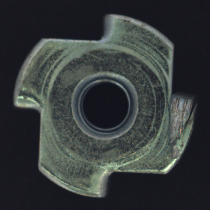} &
     	\includegraphics[width=\widthface\textwidth]{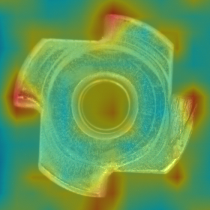} &
            \includegraphics[width=\widthface\textwidth]{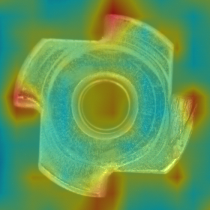} &
            \includegraphics[width=\widthface\textwidth]{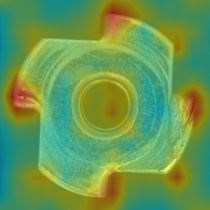} &
            \includegraphics[width=\widthface\textwidth]{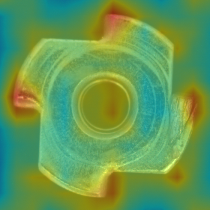} &
            \includegraphics[width=\widthface\textwidth]{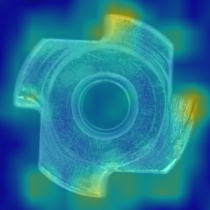} &
	    \includegraphics[width=\widthface\textwidth]{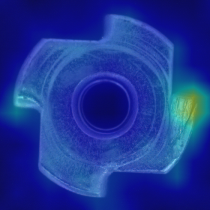} &
	    \includegraphics[width=\widthface\textwidth]{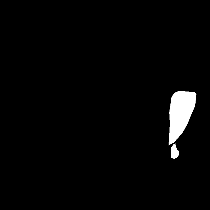} \\
     	\includegraphics[width=\widthface\textwidth]{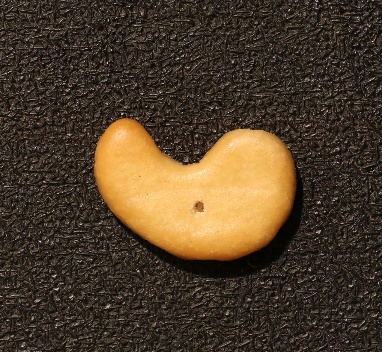} &
     	\includegraphics[width=\widthface\textwidth]{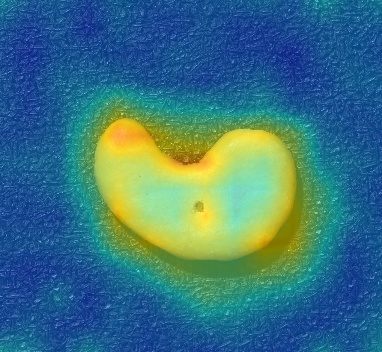} &
            \includegraphics[width=\widthface\textwidth]{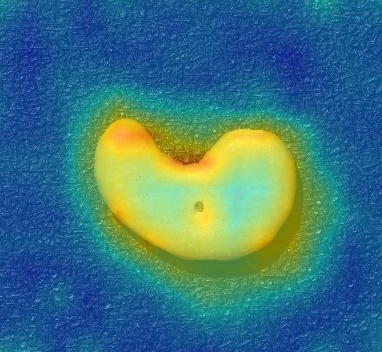} &
            \includegraphics[width=\widthface\textwidth]{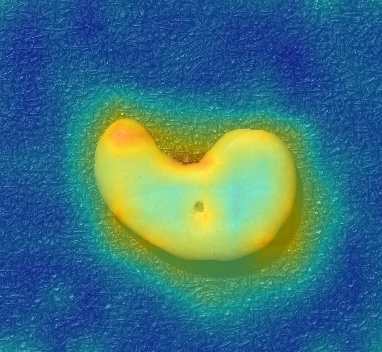} &
            \includegraphics[width=\widthface\textwidth]{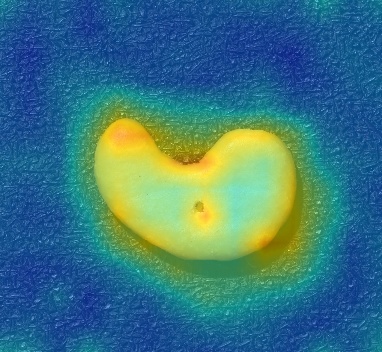} &
            \includegraphics[width=\widthface\textwidth]{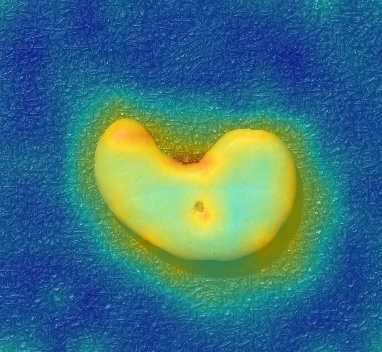} &
	    \includegraphics[width=\widthface\textwidth]{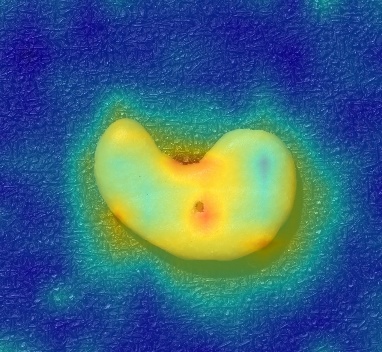} &
	    \includegraphics[width=\widthface\textwidth]{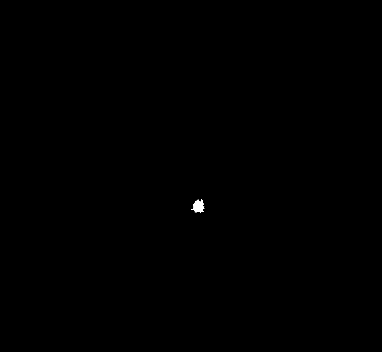} \\

            \includegraphics[width=\widthface\textwidth]{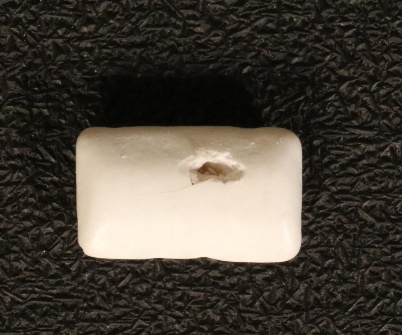} &
     	\includegraphics[width=\widthface\textwidth]{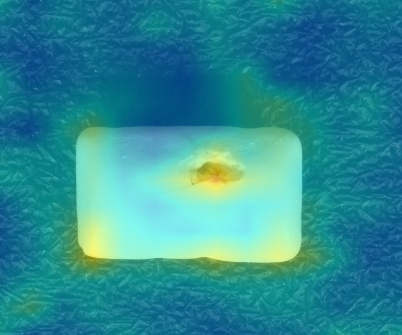} &
            \includegraphics[width=\widthface\textwidth]{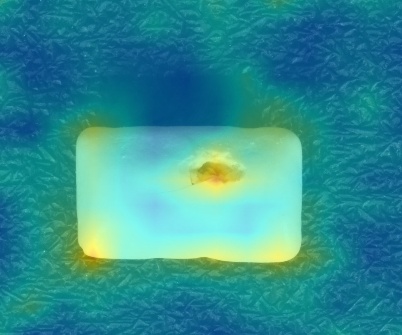} &
            \includegraphics[width=\widthface\textwidth]{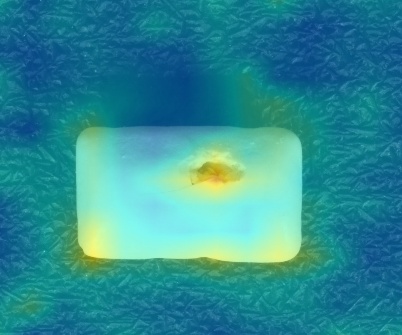} &
            \includegraphics[width=\widthface\textwidth]{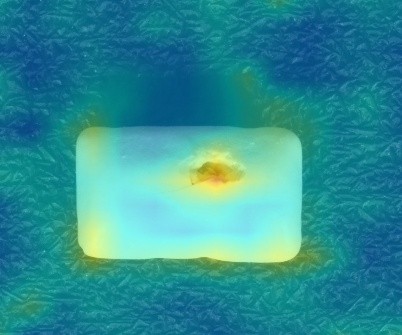} &
            \includegraphics[width=\widthface\textwidth]{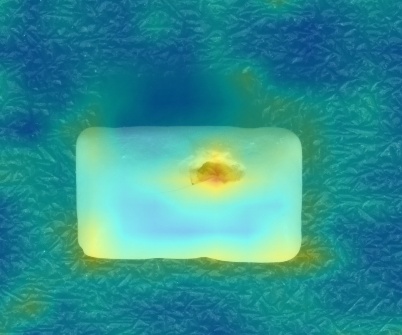} &
	    \includegraphics[width=\widthface\textwidth]{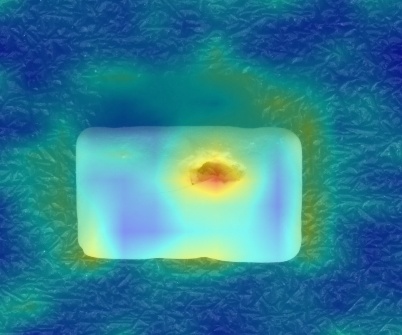} &
	    \includegraphics[width=\widthface\textwidth]{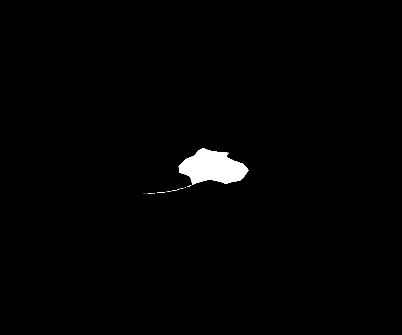} 
	\end{tabular}}
	\caption{Qualitative evaluation in MVTec-AD~\cite{bergmann2019mvtec} and VisA~\cite{zou2022spot}. The intensity of red in the heatmap indicates a higher likelihood of defects, whereas blue signifies a lower probability. Our approach outperforms existing baselines by significantly mitigating semantic feature conflict and enhancing defect localization accuracy. \color{red}(Please zoom in for more details.)}
	\label{vis}
\end{figure}



\noindent \textbf{Evaluation Metrics} Currently, there are two metrics for continual learning: average accuracy (ACC)~\cite{lopez2017gradient} and forgetting measure (FM)~\cite{chaudhry2018riemannian}. However, for individual tasks, we usually use Pixel-level AUROC, $A^{\mathbf{pix}}$, and Image-level AUROC, $A^{\mathbf{img}}$, to characterize the detection accuracy. In this task, we combine these two goals and define four metrics, as in Eq.~\eqref{metrics1},
\begin{equation} \label{metrics1}
\small
ACC = \left\{ \begin{aligned}
	&\frac{1}{N}\sum_{i=1}^{N-1}{A_{N,i}^{\mathbf{pix}}}\,\,\\
	&\frac{1}{N}\sum_{i=1}^{N-1}{A_{N,i}^{\mathbf{img}}}\\
\end{aligned} \right. , 
FM = \left\{ \begin{aligned}
	&\frac{1}{N-1}\sum_{i=1}^{N-1}{\max_{b\in \{1,\cdots ,N-1\}}}\left( A_{b,i}^{\mathbf{pix}}-A_{N,i}^{\mathbf{pix}} \right)\\
	&\frac{1}{N-1}\sum_{i=1}^{N-1}{\max_{b\in \{1,\cdots ,N-1\}}}\left( A_{b,i}^{\mathbf{img}}-A_{N,i}^{\mathbf{img}} \right)\\
\end{aligned} \right. .
\end{equation}


\subsection{Experiment Performance}
\noindent \textbf{Quantitative Evaluation} Table~\ref{perfor} (A) (B) shows that our method achieves SOTA performance with different experiment settings in pixel-level (A) and Image-level (B), respectively. We observe that the network performance is catastrophic forgetting in the reconstruction-based approach (UniAD~\cite{you2022unified}). While some continuous learning strategies~\cite{zenke2017si,kirkpatrick2017overcoming,aljundi2018mas,wang2022continual} can partially mitigate this problem, our approach provides an optimal solution. Besides, compared with CAD-based methods, our method can offer not only pixel-level location but also SOTA performance for incrementing more objects. Overall, our algorithm can overcome the above drawbacks and maintain a low level of forgetting.

\noindent \textbf{Qualitative Evaluation for Defect Localization} Fig.~\ref{vis} shows the defect location (heatmap) of our method and other SOTA reconstruction-based approaches. We follow the same strategy in UniAD~\cite{you2022unified} to calculate the heatmap. Regions, where the occurring chance of defects is higher, are colored red. Conversely, regions, where defects are almost impossible to occur, are colored blue. Our method can significantly reduce semantic feature conflicts (In Fig.~\ref{problem}) and output more accurate defect locations.

\subsection{Ablation Study}

We conduct the ablation study on MvTec AD~\cite{bergmann2019mvtec} to evaluate the effectiveness of different components in our proposed method.

\begin{figure}[t]
    \centering
	\newcommand\widthface{0.2}
	\resizebox{1.0\linewidth}{!}{
	\begin{tabular}{cccccc}
      Input& w/o OSOA  &w/o US &w/o SCL & Ours&Ground Truth \\
	    \includegraphics[width=\widthface\textwidth]{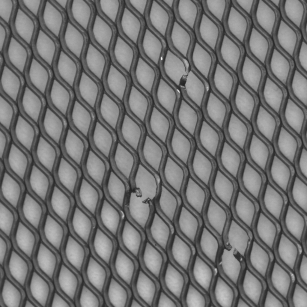} &
            \includegraphics[width=\widthface\textwidth]{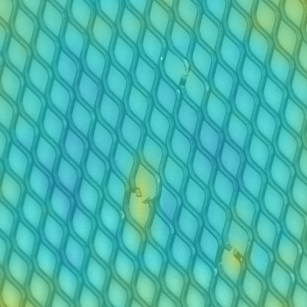} &
            \includegraphics[width=\widthface\textwidth]{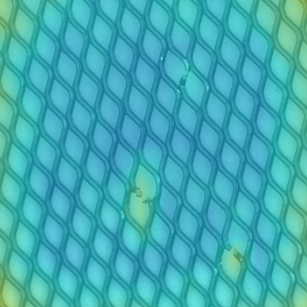} &
            \includegraphics[width=\widthface\textwidth]{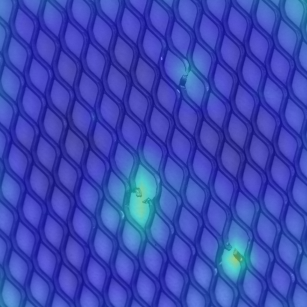} &
	    \includegraphics[width=\widthface\textwidth]{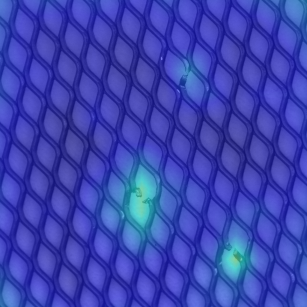} &
	    \includegraphics[width=\widthface\textwidth]{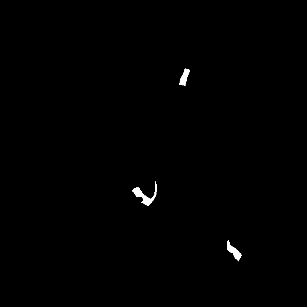}
	\end{tabular}}
    \caption{Qualitative evaluation of ablation study. Compared with ``w/o US'' and ``w/o OSOA'', it is obvious that our method can be more accurate in locating defects. Compared with ``w/o SCL'', since increasing the available space of the model, our algorithm can further reduce the feature conflict and reduce more interference in the background. \color{red}(Please zoom in for more details.)}
    \label{ab3}
\end{figure}

\noindent \textbf{Effectiveness of Object-Aware Self-Attention} To demonstrate the effectiveness of category labels, we ablate Object-Aware Self-Attention in our structure. ``w/o OASA'' in Table~\ref{ablation} shows that this structure can reduce catastrophic forgetting by category information insertion. Moreover, in Fig.~\ref{ab3} (``w/o OASA''), if there is no semantic boundary, the network can still detect anomalies in the current object, but serious feature conflicts will occur in other regions of the object, further affecting the overall performance of the network.

\begin{table*}[!t]
\centering
  \caption{Quantitative evaluation in ablation study. The results show that three components, Object-Aware Self-Attention, semantic compression loss, and the updating strategy, contribute to the performance improvement of the whole framework. \textcolor{red}{Red} and \textcolor{red!50}{Gray Red} represents the best image-level and pixel-level performance respectively.}
    \resizebox{1.0\textwidth}{!}{
    \begin{tabular}{lcc|cc|cc|cc}
    \toprule
    \multirow{2}[4]{*}{} & \multicolumn{2}{c|}{$\mathbf{14-1\ with \ 1\ Step}$} & \multicolumn{2}{c|}{$\mathbf{10-5\ with \ 1\ Step}$} & \multicolumn{2}{c|}{$\mathbf{3-3\ with \ 4\ Steps}$} & \multicolumn{2}{c}{$\mathbf{10-1\ with \ 5\ Steps}$} \\
    \cmidrule{2-9}          & ACC($\uparrow$)   & FM($\downarrow$)    & ACC($\uparrow$)   & FM($\downarrow$)    & ACC($\uparrow$)   & FM($\downarrow$)    & ACC($\uparrow$)   & FM($\downarrow$) \\

    \midrule
    w/o OASA & 89.3 / \textcolor{gray}{92.5} & 13.3 / \textcolor{gray}{8.2} &  87.6 / \textcolor{gray}{90.6} & 16.5 / \textcolor{gray}{13.5} & 81.4 / \textcolor{gray}{88.9} & 12.2 / \textcolor{gray}{10.8} & 84.2 / \textcolor{gray}{90.3} & 14.2 / \textcolor{gray}{8.6} \\
        
    w/o SCL  & 95.5 / \textcolor{gray}{96.1} & 2.4 / \textcolor{gray}{0.8} & 92.0 / \textcolor{gray}{94.3} & \textcolor{red!60}{8.1} / \textcolor{red!60}{6.0}& 82.7 / \textcolor{gray}{88.3} & 10.6 / \textcolor{gray}{11.7} & 93.4 / \textcolor{red!60}{95.1} & 3.6 / \textcolor{gray}{2.7}\\
    
    w/o US & 95.9 / \textcolor{gray}{96.2} & 1.1 / \textcolor{gray}{0.7}& 88.6 / \textcolor{gray}{93.3} & 16.5 / \textcolor{gray}{8.5} & 83.5 / \textcolor{gray}{89.5} & 10.9 / \textcolor{gray}{10.3} & 85.6 / \textcolor{gray}{90.5} & 13.5 / \textcolor{gray}{8.4} \\
    
    \midrule
    \rowcolor{class2}Ours  & \textcolor{red}{96.0} / \textcolor{red!60}{96.3} & \textcolor{red}{1.0}  / \textcolor{red!60}{0.6}& \textcolor{red}{92.2} /  \textcolor{red!60}{94.4}& 9.3  / \textcolor{gray}{6.3} & \textcolor{red}{84.2} /  \textcolor{red!60}{91.1} & 10.0 / \textcolor{red!60}{8.4}& \textcolor{red}{94.2} / \textcolor{red!60}{95.1}& \textcolor{red}{3.2} / \textcolor{red!60}{1.0}\\
    \bottomrule
    \end{tabular}}
  \label{ablation}
\end{table*}

\noindent \textbf{Effectiveness of Semantic Compression Loss} To demonstrate the effectiveness of Semantic Compression Loss, we ablate this loss in our training. ``w/o SCL'' in Table~\ref{ablation} shows that this structure can reduce catastrophic forgetting by preserving more space for new objects. Besides, in Fig.~\ref{ab3} (``w/o SCL''), by compressing more space for new objects, we can obtain clearer defect localization in detail, which indicates that more network space can be explained in incremental learning by semantic compression loss.

\noindent \textbf{Effectiveness of Our Updating Strategy} To demonstrate the effectiveness of our updating strategy, we ablate this updating method, and ``w/o US'' in Table~\ref{ablation} shows that this method can reduce catastrophic forgetting. Also, in Fig.~\ref{ab3} (``w/o US''), our algorithm can help memorize important semantic spaces for old objects, thus reducing the rewriting of semantic space in the reconstruction network, thus leading to a better localization performance.

\section{Conclusion}
To conclude, our incremental unified framework effectively integrates the multi-objects detection model with object-incremental learning, significantly enhancing the dynamic of defect inspection systems. Leveraging Object-Aware Self-Attention, Semantic Compression Loss, and updating strategy, we demarcate semantic boundaries for objects and minimize interference during incrementing new objects. Benchmarks demonstrate our SOTA performance at both the image level and pixel level. Widespread deployment of this technology will increase efficiency and reduce overhead in industrial manufacturing.

\section*{Acknowledgements} The research work is sponsored by AIR@InnoHK.

Besides, this work is funded by National Natural Science Foundation of China Grant No. 72371271, the Guangzhou Industrial Information and Intelligent Key Laboratory Project (No. 2024A03J0628), the Nansha Key Area Science and Technology Project (No. 2023ZD003), Project No. 2021JC02X191, and Natural Science Foundation of Zhejiang Pvovince, China (No. LD24F020002).


%
%
\bibliographystyle{splncs04}
\bibliography{aaai24}

\begin{thebibliography}{10}
\providecommand{\url}[1]{\texttt{#1}}
\providecommand{\urlprefix}{URL }
\providecommand{\doi}[1]{https://doi.org/#1}

\bibitem{akcay2019ganomaly}
Akcay, S., Atapour-Abarghouei, A., Breckon, T.P.: Ganomaly: Semi-supervised anomaly detection via adversarial training. In: Computer Vision--ACCV 2018: 14th Asian Conference on Computer Vision, Perth, Australia, December 2--6, 2018, Revised Selected Papers, Part III 14. pp. 622--637. Springer (2019)

\bibitem{aljundi2018mas}
Aljundi, R., Babiloni, F., Elhoseiny, M., Rohrbach, M., Tuytelaars, T.: Memory aware synapses: Learning what (not) to forget. In: Proceedings of the European conference on computer vision (ECCV). pp. 139--154 (2018)

\bibitem{bergmann2019mvtec}
Bergmann, P., Fauser, M., Sattlegger, D., Steger, C.: Mvtec ad--a comprehensive real-world dataset for unsupervised anomaly detection. In: Proceedings of the IEEE/CVF conference on computer vision and pattern recognition (CVPR). pp. 9592--9600 (2019)

\bibitem{bergmann2018improving}
Bergmann, P., L{\"o}we, S., Fauser, M., Sattlegger, D., Steger, C.: Improving unsupervised defect segmentation by applying structural similarity to autoencoders. arXiv preprint arXiv:1807.02011  (2018)

\bibitem{chang2022tire}
Chang, C.Y., Su, Y.D., Li, W.Y.: Tire bubble defect detection using incremental learning. Applied Sciences  \textbf{12}(23),  12186 (2022)

\bibitem{chaudhry2018riemannian}
Chaudhry, A., Dokania, P.K., Ajanthan, T., Torr, P.H.: Riemannian walk for incremental learning: Understanding forgetting and intransigence. In: Proceedings of the European conference on computer vision (ECCV). pp. 532--547 (2018)

\bibitem{chen2021defect}
Chen, C.H., Tu, C.H., Li, J.D., Chen, C.S.: Defect detection using deep lifelong learning. In: 2021 IEEE 19th International Conference on Industrial Informatics (INDIN). pp.~1--6. IEEE (2021)

\bibitem{chen2022utrad}
Chen, L., You, Z., Zhang, N., Xi, J., Le, X.: Utrad: Anomaly detection and localization with u-transformer. Neural Networks  \textbf{147},  53--62 (2022)

\bibitem{cohen2020sub}
Cohen, N., Hoshen, Y.: Sub-image anomaly detection with deep pyramid correspondences. arXiv preprint arXiv:2005.02357  (2020)

\bibitem{defard2021padim}
Defard, T., Setkov, A., Loesch, A., Audigier, R.: Padim: a patch distribution modeling framework for anomaly detection and localization. In: International Conference on Pattern Recognition. pp. 475--489. Springer (2021)

\bibitem{Ding_2021_ICCV}
Ding, X., Hao, T., Tan, J., Liu, J., Han, J., Guo, Y., Ding, G.: Resrep: Lossless cnn pruning via decoupling remembering and forgetting. In: Proceedings of the IEEE/CVF International Conference on Computer Vision (ICCV). pp. 4510--4520 (October 2021)

\bibitem{han2023svdiff}
Han, L., Li, Y., Zhang, H., Milanfar, P., Metaxas, D., Yang, F.: Svdiff: Compact parameter space for diffusion fine-tuning. arXiv preprint arXiv:2303.11305  (2023)

\bibitem{hoefler2021sparsity}
Hoefler, T., Alistarh, D., Ben-Nun, T., Dryden, N., Peste, A.: Sparsity in deep learning: Pruning and growth for efficient inference and training in neural networks. The Journal of Machine Learning Research  \textbf{22}(1),  10882--11005 (2021)

\bibitem{huong2022federated}
Huong, T.T., Bac, T.P., Ha, K.N., Hoang, N.V., Hoang, N.X., Hung, N.T., Tran, K.P.: Federated learning-based explainable anomaly detection for industrial control systems. IEEE Access  \textbf{10},  53854--53872 (2022)

\bibitem{jayasekara2023detecting}
Jayasekara, H., Zhang, Q., Yuen, C., Zhang, M., Woo, C.W., Low, J.: Detecting anomalous solder joints in multi-sliced pcb x-ray images: A deep learning based approach. SN Computer Science  \textbf{4}(3), ~307 (2023)

\bibitem{kahler2022anomaly}
K{\"a}hler, F., Schmedemann, O., Sch{\"u}ppstuhl, T.: Anomaly detection for industrial surface inspection: Application in maintenance of aircraft components. Procedia CIRP  \textbf{107},  246--251 (2022)

\bibitem{khalil2022efficient}
Khalil, A.A., E~Ibrahim, F., Abbass, M.Y., Haggag, N., Mahrous, Y., Sedik, A., Elsherbeeny, Z., Khalaf, A.A., Rihan, M., El-Shafai, W., et~al.: Efficient anomaly detection from medical signals and images with convolutional neural networks for internet of medical things (iomt) systems. International Journal for Numerical Methods in Biomedical Engineering  \textbf{38}(1),  e3530 (2022)

\bibitem{kirkpatrick2017overcoming}
Kirkpatrick, J., Pascanu, R., Rabinowitz, N., Veness, J., Desjardins, G., Rusu, A.A., Milan, K., Quan, J., Ramalho, T., Grabska-Barwinska, A., et~al.: Overcoming catastrophic forgetting in neural networks. Proceedings of the national academy of sciences  \textbf{114}(13),  3521--3526 (2017)

\bibitem{Li_2021_cut}
Li, C.L., Sohn, K., Yoon, J., Pfister, T.: Cutpaste: Self-supervised learning for anomaly detection and localization. In: Proceedings of the IEEE/CVF Conference on Computer Vision and Pattern Recognition (CVPR). pp. 9664--9674 (June 2021)

\bibitem{li2022towards}
Li, W., Zhan, J., Wang, J., Xia, B., Gao, B.B., Liu, J., Wang, C., Zheng, F.: Towards continual adaptation in industrial anomaly detection. In: Proceedings of the 30th ACM International Conference on Multimedia. pp. 2871--2880 (2022)

\bibitem{lin2020hrank}
Lin, M., Ji, R., Wang, Y., Zhang, Y., Zhang, B., Tian, Y., Shao, L.: Hrank: Filter pruning using high-rank feature map. In: Proceedings of the IEEE/CVF conference on computer vision and pattern recognition (CVPR). pp. 1529--1538 (2020)

\bibitem{liu2024unsupervised}
Liu, J., Wu, K., Nie, Q., Chen, Y., Gao, B.B., Liu, Y., Wang, J., Wang, C., Zheng, F.: Unsupervised continual anomaly detection with contrastively-learned prompt. arXiv preprint arXiv:2401.01010  (2024)

\bibitem{liu2020towards}
Liu, W., Li, R., Zheng, M., Karanam, S., Wu, Z., Bhanu, B., Radke, R.J., Camps, O.: Towards visually explaining variational autoencoders. In: Proceedings of the IEEE/CVF Conference on Computer Vision and Pattern Recognition. pp. 8642--8651 (2020)

\bibitem{liu2023simplenet}
Liu, Z., Zhou, Y., Xu, Y., Wang, Z.: Simplenet: A simple network for image anomaly detection and localization. In: Proceedings of the IEEE/CVF Conference on Computer Vision and Pattern Recognition. pp. 20402--20411 (2023)

\bibitem{lopez2017gradient}
Lopez-Paz, D., Ranzato, M.: Gradient episodic memory for continual learning. Advances in neural information processing systems  \textbf{30} (2017)

\bibitem{lu2022deep}
Lu, B., Xu, D., Huang, B.: Deep-learning-based anomaly detection for lace defect inspection employing videos in production line. Advanced Engineering Informatics  \textbf{51},  101471 (2022)

\bibitem{9340601}
Peng, C., Zhang, K., Ma, Y., Ma, J.: Cross fusion net: A fast semantic segmentation network for small-scale semantic information capturing in aerial scenes. IEEE Transactions on Geoscience and Remote Sensing  \textbf{60},  1--13 (2022). \doi{10.1109/TGRS.2021.3053062}

\bibitem{pirnay2022inpainting}
Pirnay, J., Chai, K.: Inpainting transformer for anomaly detection. In: International Conference on Image Analysis and Processing. pp. 394--406. Springer (2022)

\bibitem{reiss2021panda}
Reiss, T., Cohen, N., Bergman, L., Hoshen, Y.: Panda: Adapting pretrained features for anomaly detection and segmentation. In: Proceedings of the IEEE/CVF Conference on Computer Vision and Pattern Recognition (CVPR). pp. 2806--2814 (2021)

\bibitem{roth2022towards}
Roth, K., Pemula, L., Zepeda, J., Sch{\"o}lkopf, B., Brox, T., Gehler, P.: Towards total recall in industrial anomaly detection. In: Proceedings of the IEEE/CVF Conference on Computer Vision and Pattern Recognition. pp. 14318--14328 (2022)

\bibitem{sun2022new}
Sun, C., Gao, L., Li, X., Gao, Y.: A new knowledge distillation network for incremental few-shot surface defect detection. arXiv preprint arXiv:2209.00519  (2022)

\bibitem{sun2023continual}
Sun, W., Al~Kontar, R., Jin, J., Chang, T.S.: A continual learning framework for adaptive defect classification and inspection. Journal of Quality Technology pp. 1--17 (2023)

\bibitem{towill1997analysis}
Towill, D.R., Evans, G.N., Cheema, P.: Analysis and design of an adaptive minimum reasonable inventory control system. Production Planning \& Control  \textbf{8}(6),  545--557 (1997)

\bibitem{wang2022continual}
Wang, Z., Liu, L., Duan, Y., Kong, Y., Tao, D.: Continual learning with lifelong vision transformer. In: Proceedings of the IEEE/CVF Conference on Computer Vision and Pattern Recognition (CVPR). pp. 171--181 (2022)

\bibitem{xie2023pushing}
Xie, G., Wang, J., Liu, J., Zheng, F., Jin, Y.: Pushing the limits of fewshot anomaly detection in industry vision: Graphcore. arXiv preprint arXiv:2301.12082  (2023)

\bibitem{yang2024defectspectrumgranularlook}
Yang, S., Chen, Z., Chen, P., Fang, X., Liu, S., Chen, Y.: Defect spectrum: A granular look of large-scale defect datasets with rich semantics (2024), \url{https://arxiv.org/abs/2310.17316}

\bibitem{yildiz2022automated}
Yildiz, O., Chan, H., Raghavan, K., Judge, W., Cherukara, M.J., Balaprakash, P., Sankaranarayanan, S., Peterka, T.: Automated continual learning of defect identification in coherent diffraction imaging. In: 2022 IEEE/ACM International Workshop on Artificial Intelligence and Machine Learning for Scientific Applications (AI4S). pp.~1--6. IEEE (2022)

\bibitem{you2022unified}
You, Z., Cui, L., Shen, Y., Yang, K., Lu, X., Zheng, Y., Le, X.: A unified model for multi-class anomaly detection. Advances in Neural Information Processing Systems  \textbf{35},  4571--4584 (2022)

\bibitem{Zavrtanik2021DRMA}
Zavrtanik, V., Kristan, M., Skovcaj, D.: Draem - a discriminatively trained reconstruction embedding for surface anomaly detection. 2021 IEEE/CVF International Conference on Computer Vision (ICCV)  (2021)

\bibitem{zenke2017si}
Zenke, F., Poole, B., Ganguli, S.: Continual learning through synaptic intelligence. In: International conference on machine learning. pp. 3987--3995. PMLR (2017)

\bibitem{zhao2023omnial}
Zhao, Y.: Omnial: A unified cnn framework for unsupervised anomaly localization. In: Proceedings of the IEEE/CVF Conference on Computer Vision and Pattern Recognition. pp. 3924--3933 (2023)

\bibitem{zhou2020siamese}
Zhou, X., Liang, W., Shimizu, S., Ma, J., Jin, Q.: Siamese neural network based few-shot learning for anomaly detection in industrial cyber-physical systems. IEEE Transactions on Industrial Informatics  \textbf{17}(8),  5790--5798 (2020)

\bibitem{zou2022spot}
Zou, Y., Jeong, J., Pemula, L., Zhang, D., Dabeer, O.: Spot-the-difference self-supervised pre-training for anomaly detection and segmentation. In: European Conference on Computer Vision. pp. 392--408. Springer (2022)

\end{thebibliography}
\end{document}